\documentclass[10pt,twocolumn,letterpaper]{article}

\usepackage{wacv}
\usepackage{times}
\usepackage{epsfig}
\usepackage{graphicx}
\usepackage{amsmath,amssymb} 
\usepackage{color}
\usepackage{wrapfig}
\usepackage{sidecap}
\makeatletter
\@namedef{ver@everyshi.sty}{}
\makeatother
\usepackage{tikz}
\usepackage{epsfig}
\usepackage{tabularx}
\usepackage{booktabs}
\usepackage{bbm}
\usepackage{mathtools}
\usepackage{epstopdf}
\usepackage{subcaption}
\usepackage{pgfplots} 
\usepackage{pgfplotstable} 

\pgfplotsset{compat=newest} 
\usetikzlibrary{plotmarks} 
\usetikzlibrary{colorbrewer} 

\usepackage[pagebackref=true,breaklinks=true,letterpaper=true,colorlinks,bookmarks=false]{hyperref}

\newcommand{\Alina}[1]{}
\newcommand{\vitto}[1]{}

\newcommand{\mypar}[1]{\vspace{0.5em}{\noindent\textbf{#1}}}

\def\wacvPaperID{740} 

\wacvfinalcopy 
\ifwacvfinal
\def\assignedStartPage{9876} 
\fi

\ifwacvfinal
\usepackage[breaklinks=true,bookmarks=false]{hyperref}
\else
\usepackage[pagebackref=true,breaklinks=true,colorlinks,bookmarks=false]{hyperref}
\fi

\ifwacvfinal
\else
\fi

\begin{document}

\title{Efficient video annotation with visual interpolation and frame selection guidance}

\author{Alina Kuznetsova \qquad 
Aakrati Talati 
\qquad  Yiwen Luo 
\qquad  Keith Simmons
\qquad  Vittorio Ferrari \\
\\
Google Research \\
{\tt\small \{akuznetsa,aakrati,yiwenluo,eskiman,vittoferrari\}@google.com}} 

\maketitle

\begin{abstract}
We introduce a unified framework for generic video annotation with bounding boxes. 
Video annotation is a long-standing problem, as it is a tedious and time-consuming process.
We tackle two important challenges of video annotation: (1) automatic temporal interpolation and extrapolation of bounding boxes provided by a human annotator on a subset of all frames,
and (2) automatic selection of frames to annotate manually.
Our contribution is two-fold: first, we propose a model that has both interpolating and extrapolating capabilities; second, we propose a guiding mechanism that sequentially generates suggestions for what frame to annotate next, based on the annotations made previously.
We extensively evaluate our approach on several challenging datasets in simulation and demonstrate a reduction in terms of the number of manual bounding boxes drawn by $60\%$ over linear interpolation and by $35\%$ over an off-the-shelf tracker. Moreover, we also show $10\%$ annotation time improvement over a state-of-the-art method for video annotation with bounding boxes~\cite{pathtrack}. 
Finally, we run human annotation experiments and provide extensive analysis of the results, showing that our approach reduces actual measured annotation time by $50\%$ compared to commonly used linear interpolation.
\end{abstract}

\section{Introduction}\label{sec:intro}
\begin{figure}[t]
\centering
\includegraphics[width=0.98\linewidth]{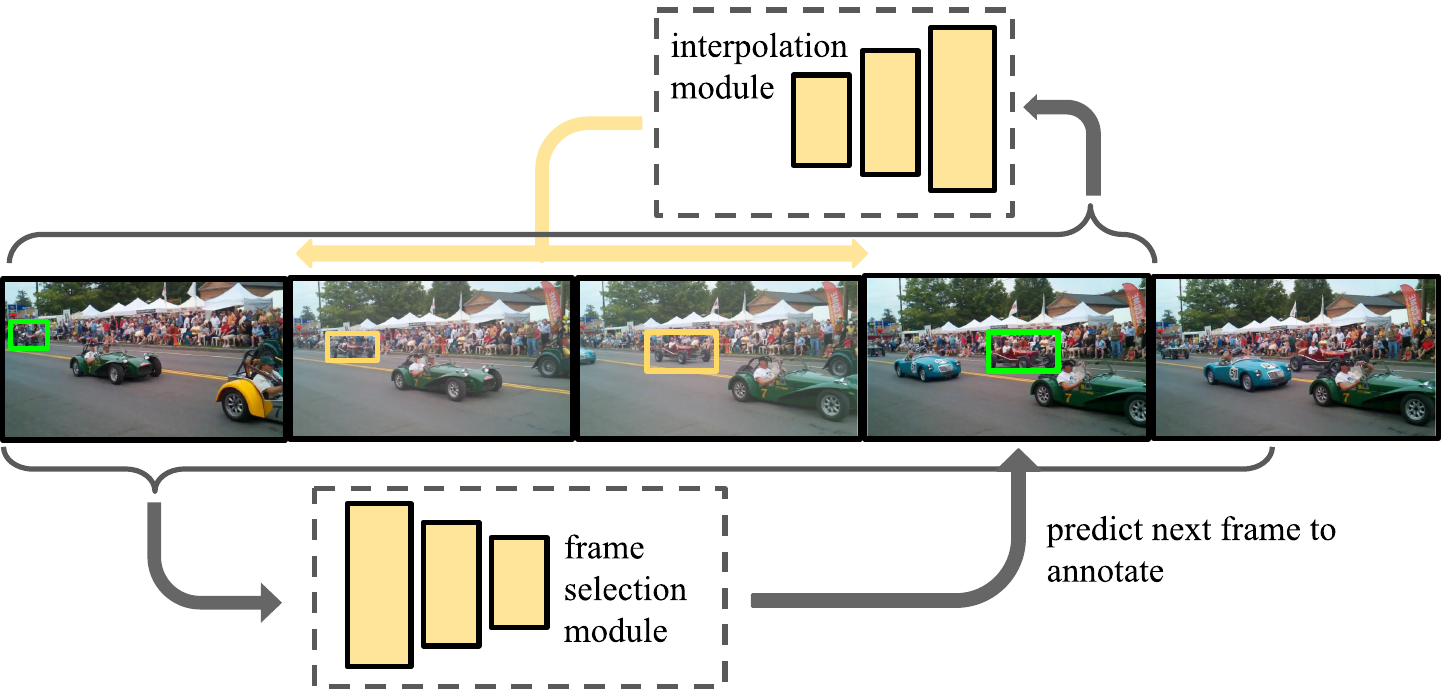}
\vspace{-0.2cm}
\caption{\small
\vitto{I believe the figure is not quite right. The interpolation module inputs two boxes, not just one, right? There should be a green arrow from the first frame to the interpolation module}
Overview of our video annotation process. A human annotator draws a box on the first frame of the video; then our guiding frame selection mechanism predicts the next frame to annotate and the process iterates. Our method automatically and accurately interpolates bounding boxes for all frames that were not directly annotated by the human. Hence, at the end of the process object annotations are generated for all frames.\Alina{updated}}\label{fig:teaser}
\vspace{-0.4cm}
\end{figure}

Progress in machine learning techniques depends on the availability of large volumes of high quality annotated data.
Recently several large scale image datasets have appeared~\cite{openimagesv4,objects365,lvis}, as well as large-scale tracking benchmarks~\cite{got10k,lasot}, but they required tremendous annotation resources to create~\cite{openimagesv4,vatic}.
The reported annotation time for box annotation ranges between $5.2$~\cite{pathtrack} and $20$~\cite{ImageNetpaper} seconds per bounding box.
Hence, the time to create a dataset of similar size to Got10k~\cite{got10k} requires about $3000$ - $8000$ hours of work just for the box annotation stage (provided each box is annotated individually).
Due to this high cost, none of the existing large-scale video benchmarks provides exhaustive annotations, not even at the video clip level.
Going beyond bounding boxes, video instance segmentation datasets are even smaller~\cite{youtbevos,davis}. 
Being able to easily develop such datasets would speed up the progress in unconstrained video understanding~\cite{ava,got10k}.
In this paper we propose an efficient video annotation. 
Our framework consists of two interacting modules:
(1) a module for interpolation and extrapolation of annotations created by a human annotator (we call it visual interpolation below for simplicity)
and (2) a guiding mechanism that selects which frame to annotate.
During the annotation process, a human annotator starts by annotating the object in a single frame.
The guiding mechanism produces a prediction for which frame to annotate next and the visual interpolation module propagates the annotation to other frames.
Note, that unlike traditional active learning approaches~\cite{vatical,alprop} the guiding mechanism produces frame proposals in a sequential manner and per track. 
See Fig~\ref{fig:teaser} for an overview of the process.

Single-object tracking techniques made big progress in recent years~\cite{Kristan2019a}. 
In particular siamese trackers~\cite{siamesefc,siameserpn,siammask} showed excellent results on tracking benchmarks. 
Moreover, those models offer real-time performance, making them suitable for an interactive annotation process. 
However those techniques are underexplored for annotation purposes.
One reason is the lack of a track correction mechanism
that would allow to efficiently correct the output of the tracker.
Here we propose to alleviate this drawback by extending a siamese tracker to enable corrections and to take advantage of ground-truth annotations in multiple frames, which become available during the annotation process.
Our guiding mechanism is based on the observation that not all frames are equally useful for annotation. For example, a frame where an object is heavily occluded is unlikely to allow the visual interpolation module to propagate well to other frames. Hence, we propose to rank unannotated frames based on the expected quality of annotations generated by our visual interpolation module if those frames would be selected for annotation.
The ranking is based on pairwise comparisons of the candidate unannotated frames. In this fashion, our two proposed modules interact and are part of an integrated system.
In summary, we propose: (1) a visual interpolation module that adapts existing trackers to the annotation scenario; (2) a guiding module that automatically selects frames to send for annotation;
(3) an integrated framework where both modules work smoothly together.
We highlight that the proposed framework allows a real interactive annotation process, as it does not require offline pre- or post-processing.
We provide extensive experimental ablation studies on the ImageNetVID dataset~\cite{imagenetvid}.
We compare our approach to the traditionally used linear interpolation and forward tracking using the same base siamese model. Our approach reduces by $60\%$ the number of manually drawn boxes compared to linear interpolation, and by $35\%$ compared to tracking at a fixed quality ( $80\%$ of all frames annotated at IoU $>0.7$ ).
Next, we perform experiments with real human annotators on the Got10k~\cite{got10k} dataset and show that our framework allows to reduce actual annotation time by $50\%$ compared to annotation time when using linear interpolation.
Finally, we show that our framework is efficient for annotation of the challenging multi-object tracking dataset MOT2015~\cite{mot15}.
We show $10\%$ time reduction compared to the state-of-the-art framework~\cite{pathtrack} at the same level of the annotation quality. 

\section{Related Work}
\mypar{Video datasets.}
Creating video datasets with detailed localized annotations is very time-consuming and hence large-scale datasets are rare.
Recently several object tracking datasets have been proposed~\cite{trackingnet,got10k,lasot,oxuva}. While offering object diversity, they however do not contain annotations for more than a single object track per video.\footnote{\cite{oxuva} dataset does offer $13$ videos out of $185$ that contain $2-3$ objects.}
Currently only the Waymo Open Dataset~\cite{waymoopendataset} contains exhaustive annotations for all object tracks in each video. However, that dataset focuses on driving scenes and therefore has limited number of annotated classes.
The place for a large scale general purpose video dataset is still vacant and efficient video annotation methods are required to create those.

\mypar{Video annotation.}
\vitto{Alina to rewrite this}\Alina{done}
Early works on video annotation propose to speed up annotation process using geometric interpolation of annotated bounding boxes and polygons~\cite{labelme2016} across frames.
Employing video content to assist bounding boxes for video annotation was investigated in~\cite{vatic}, where the authors interpolate annotations by solving a dynamic programming problem after each new bounding box provided by a human annotator.
Several published approaches~\cite{supervoxelprop,supertrajectory} for segmentation propagation are not directly targeting the video annotation use-case and do not allow for online corrections. 
More recent work~\cite{scribblebox} proposes a solution for interactive video object segmentation annotation problem: they first obtain bounding boxes of the objects by forward tracking and subsequent curve fitting, and employ SiamMask~\cite{siammask} and scribbles to derive segmentation from box tracks. However, the initial problem of bounding box annotations remains not well studied. \cite{got10k} mentions using tracking to propagate bounding boxes between manual annotations without any further details.
A separate line of works explores training models with a small set of sparse manually annotated bounding boxes and large set of automatically labeled ones obtained via tracking~\cite{cvpr2015,eodh}. Those approaches, however, are model-specific and are not focusing on obtaining a large set of annotated data that could be re-used for training multiple models.
Finally, Pathtrack~\cite{pathtrack} proposes an approach in between the semi-supervised approaches mentioned above and manual labelling approaches like~\cite{labelme2016}, specifically tackling annotation of crowded videos. Annotators first track the center of each person with a mouse pointer through the video. Those point tracks are used to build full bounding box tracks by integrating automatic detections from a person detector.
One of the advantages of the our method over previous work is that it operates in real-time and does not require any offline pre- or post- processing. Once the infrastructure is set up, live annotation can be run immediately on new videos.

\mypar{Single-object tracking.}
Single-object tracking is a long-standing computer vision problem. The first few successful approaches~\cite{KCFs,meanshift,CONDENSATION} relied on hand-crafted features.
Recently, trackers based on deep-learned architectures~\cite{siamesefc,siamrpn,siammask,mbmd,ladcf,CFWCR,DimP} emerged in this area. 
Trackers based on Siamese architectures~\cite{siamesefc,siamrpn,DaSiamRPN,siammask} are particularly interesting, as they showed strong results on various benchmarks and are relatively simple. In our work we extend the basic model of~\cite{siamesefc,DaSiamRPN} to form our visual interpolation module.
\mypar{Active learning and other related works.}
\vitto{vitto pass done. Rewritten some small pieces here for clarity.}
It was noticed~\cite{vatical, eccv12ALGrauman} \vitto{by whom? Either we put a citation to back up this claim, or we change the text to directly start with 'One of the factors ...'} that one of the factors slowing down the annotation process is selecting frames for manual annotation and so some works explored the problem of optimal frames selection (both for video segmentation~\cite{eccv12ALGrauman} and bounding box annotation~\cite{vatical}).
However, those approaches require expensive pre-processing of all frames or online retraining of the propagation algorithm during the annotation process.
Further, the annotators have to spend time on context switching, since frames are not presented chronologically~\cite{pathtrack}. Instead, our proposed method selects frames chronologically.
Another work related to ours is BubbleNets~\cite{bubblenets}, in the domain of video instance segmentation. The task is to automatically segment an object in every frame of a video, given the ground-truth segmentation in one particular frame.
The authors show that the quality produced by a segmentation model heavily depends on which frame is given with ground-truth segmentation (which is used for fine-tuning the model).
We extend their results by investigating a more complex setting: bounding box annotation for challenging datasets containing multiple objects per frame, as opposed to focusing on a single main object per frame.
To achieve that we introduce an attention mechanism that allows the model to focus on a specific object (Sec~\ref{sec:guidance_module}). \vitto{point to section}.
Finally, different from general active learning, we do not focus on training the best quality models, but rather on annotating data in the most efficient way. This data can then be used to train any model (also beyond the particular tracker used to assist during annotation).
Our framework also does not assume any online training, which makes it more suitable for the specific scenario of interactive real-time video annotation.

\section{Video annotation framework}

Our overall framework is presented in Fig~\ref{fig:teaser}. It consists of two components: the visual interpolation module and the frame selection guiding module. 
The annotation process alternates between two steps: the human annotator drawing a bounding box in one frame and the machine carrying out the  box interpolation/extrapolation and selecting the next frame to annotate.
As we show experimentally, such human-machine collaboration is very beneficial as it reduces the total human annotation time (see Sec~\ref{sec:rater-experiments}).

\subsection{Visual interpolation}
\label{sec:visual-interpolation}

Video annotation is a time-consuming and tedious process~\cite{vatic}.
Existing approaches use linear interpolation of box geometry~\cite{labelme2016} or more complicated geometric modeling~\cite{geomboxint} that nevertheless does not rely on visual signals. 
On the other end of the spectrum are the approaches relying on visual signal only~\cite{vatic}. 
However, recent developments in single-object tracking are so far under-explored for the task of video annotation, perhaps because trackers typically assume a single target object appearance as input and do not allow any corrections after the tracking started.
To this end we propose a set of interpolation models that are based on contemporary trackers. Our model exploit visual information from multiple annotated frames at the same time, and allow to introduce and propagate corrections during the annotation process. 
Many state-of-the-art single object trackers rely on siamese architecture~\cite{siamesefc,siamrpn,siammask,DaSiamRPN}, where a single backbone is used to extract the features from the annotated frame and the subsequent video frames to combine those features in various ways to localized the target object.
We propose a simple change to siamese architectures to incorporate tracking target appearance in multiple annotated frames. This extends siamese type trackers to interpolation and allows efficient track correction mechanism. 
\begin{figure}
\centering
\includegraphics[width=0.7\linewidth]{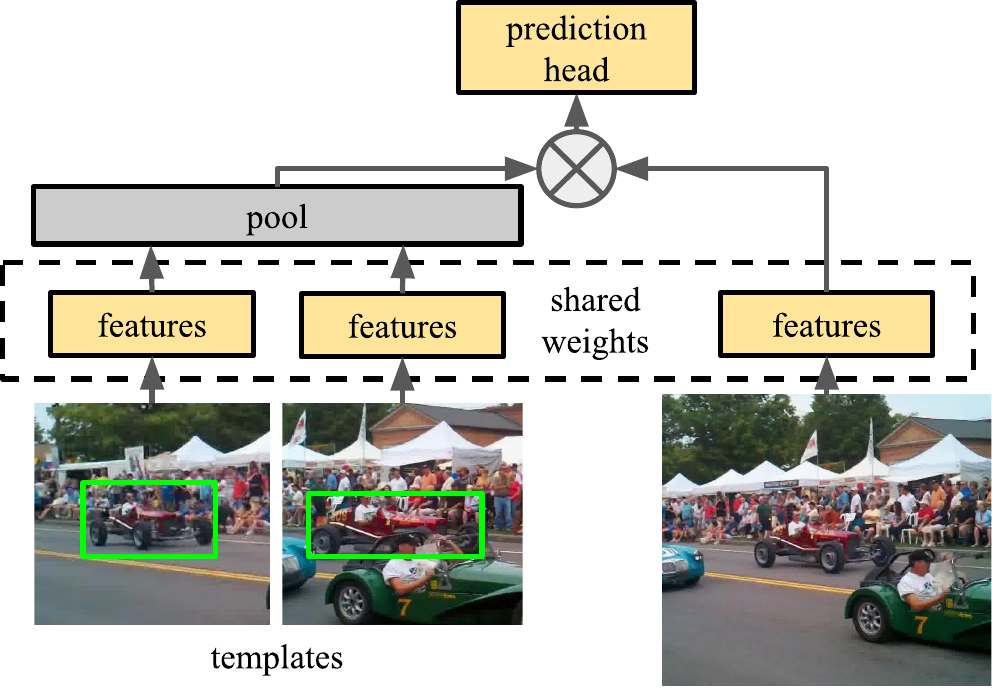}
\vspace{-0.2cm}
\caption{\small Visual interpolation model: features are first extracted from multiple templates and a joint feature vector is formed by maxpooling. The joint features are then used to derive a final prediction for an unannotated frame by convolving them with features extracted from the search space for that frame.}\label{fig:model}
\vspace{-0.4cm}
\end{figure}
In the subsequent sections we explain the proposed modification on the example of two models, SiamFC\cite{siamesefc} and DaSiamRPN~\cite{DaSiamRPN}, and in the experimental section we demonstrate that it brings significant performance improvements.

\mypar{Siamese tracking models.} The Siamese tracker model consists of two feature extractor branches with shared weights $\varphi(\cdot)$. One of the branches extracts features from the image patch containing the tracking target $z$ in the initial frame, defined by a manually annotated bounding box (we call this patch \textit{template}).
The other branch receives an image patch from the current frame $x$ (we call this patch \textit{search space}). The features extracted from the template $\varphi(z)$ are convolved with the the search space features $\varphi(x)$ to derive the score map (in case of SiamFC) or box prediction and tracker score (in case of DaSiamRPN):
\begin{equation}
  A(z, x) = \varphi(z) * \varphi(x),
\end{equation}
where $*$ denotes convolution.
During tracking, the template is obtained by cropping an image around the initial ground truth bounding box with equal width and height of $\sqrt{(w+2p)(h+2p)}$, centered around the box center and re-scaled to $127\times127$ pixels (here $w,h$ are width and height of the initial box and $p=(w+h)/4$). The search space image patch is obtained by cropping a large square patch around the current position of the target. The search space crops are computed at multiple scales for the SiamFC tracker and for a single scale for DaSiamRPN tracker.

\mypar{Visual interpolation network.} Provided ground truth annotations for the same object in multiple frames, we investigate a modification of the base siamese network to incorporate the additional visual information coming from them  (Fig~\ref{fig:model}). Let $\{z_i\}_{i=1}^{K}$ be several templates obtained for the same target in multiple frames (we call them \textit{keyframes}).
The model consists of $K+1$ feature extractors with shared weights; the features are combined by max-pooling $g(\cdot)$ as in~\cite{qi2017pointnet}. Afterwards, max-pooled features are convolved with the search space features as in the base model:
 \begin{align}
  A(z_1,\dots,z_{K},x) = g(\varphi(z_1),\dots,\varphi(z_{K})) * \varphi(x).
 \end{align}
Note, that this architecture is able to take into account arbitrary number of templates both at train and test time, potentially improving performance.

\mypar{Geometric model.} Geometric modelling for annotation propagation has an advantage over visual methods as it is robust against occlusions and bad image quality (such as blur and video decoding artifacts). Hence it is more reliable in the vicinity of the frames that contain annotations. 
To benefit from it, we blend the prediction of the visual interpolator model with a geometric interpolation model at each frame. Geometric model prediction is more reliable in a temporal neighborhood of the keyframes and less reliable further away in time. Visual interpolation generally works better for such temporally distant frames, as it follows the object visually.
To model this we introduce weight $w(\delta_t, \Delta)$, where $\delta_t$ is (absolute) offset in time to the closest keyframe and $\Delta$ is a parameter. The higher the weight $w(\delta_t, \Delta)$, the closer the overall process is to geometric interpolation model output:
\begin{align}
 w(\delta_t, \Delta) = 
\begin{cases}
0, &\delta_t > \Delta \\
\delta_t^2\Delta^{-2} - 2\delta_t\Delta^{-1}+1, &\delta_t \le \Delta
\end{cases}
\end{align}
As a geometric interpolation model we use linear interpolation between boxes in two frames. The dimensions of a box and its center position are interpolated separately.
Outside of the temporal neighborhood $(-\Delta, \Delta)$ of an annotated frame geometric interpolation has no effect. 

\mypar{Training.} We train SiamFC visual interpolation model using the train set of ImageNet VID~\cite{imagenetvid} for $10$ epochs with batch size $32$ and using momentum optimizer~\cite{momentum} with initial learning rate of $1e-3$ and exponential decay.
For DaSiamRPN we use ImageNet VID~\cite{imagenetvid}, YouTube Bounding Boxes~\cite{ytbb} and MSCOCO~\cite{mscoco} for traning as proposed in~\cite{DaSiamRPN} and using the same parameters as for SiamFC visual interpolation training. 
\begin{figure}
\centering
\includegraphics[width=0.89\linewidth]{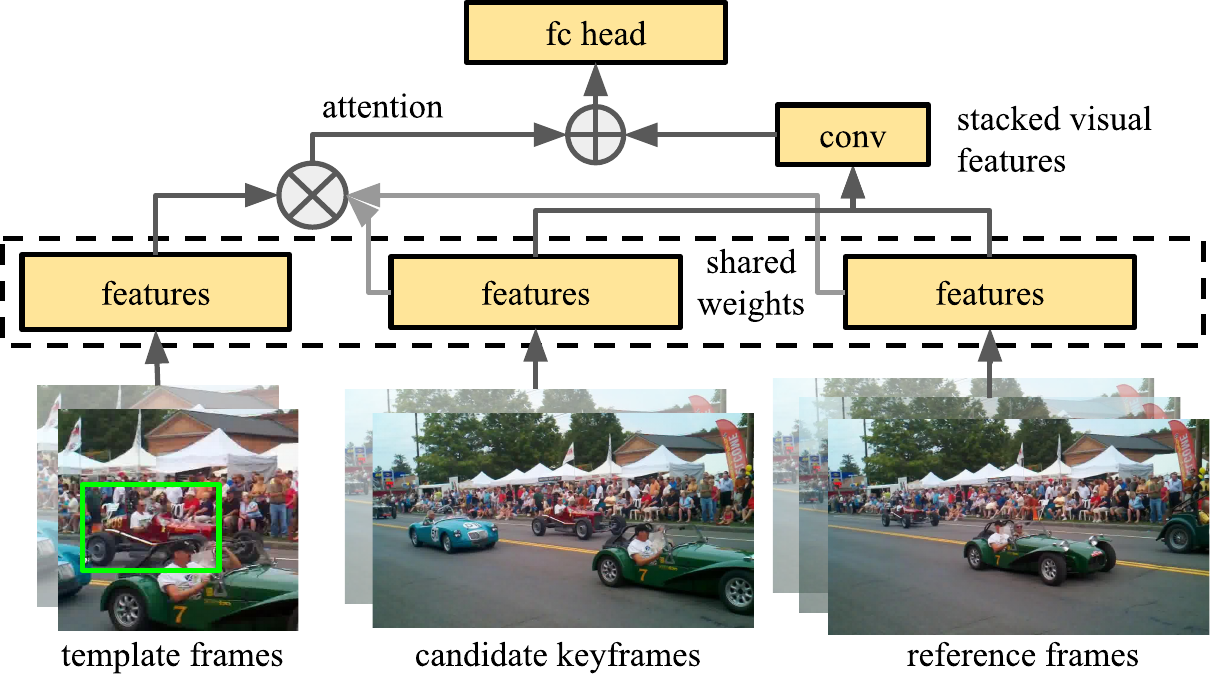}
\vspace{-0.2cm}
\caption{\small The ranking model architecture. There are three types of input:
(1) templates obtained from previously annotated frames (cropped\vitto{add box to figure under 'template frames'});
(2) two candidate keyframes\vitto{use a different candidate keyframe than a reference frame};
(3) the video representation as $N$ reference frames randomly subsampled from the video. We build an attention map on the target object by convolving the template features with the full frame features (of either the candidate keyframes or the reference frames). Then we add this attention maps to the visual features extracted from the full frames.
\vspace{-0.3cm}}\label{fig:guidance}
\end{figure}

\begin{figure}
\resizebox{\linewidth}{!}{%
\begin{tikzpicture}
\begin{axis}[width=0.89\linewidth,height=0.45\linewidth,
              ylabel=recall@0.7,
              ylabel shift={-5pt},
              xlabel=second keyframe relative index,
              xlabel shift={-3pt},
              ytick={0.2,0.4,...,1.0},
              xmin=0,
              xmax=40,
              ymin=0.0,
              ymax=1.1,
              enlargelimits=false,
              grid=both,
              grid style=densely dotted,
              legend pos = north west,
              minor grid style={white!85!black},
              major grid style={white!60!black},
              extra tick style={white!85!black},
              label style={font=\small}
              ]
  \addplot+[cyan,solid,mark=*, mark size=2, line width=1pt, mark options={fill=cyan}] table[x=Time,y=Track0] {data/TUD-Stadtmitte_135.txt};
  \addplot+[violet,dashed,mark=*,mark size=2, line width=1pt, mark options={fill=violet}] table[x=Time,y=Track1] {data/TUD-Stadtmitte_135.txt};
  \addplot+[blue,solid,mark=*, mark size=2, line width=1pt, mark options={fill=blue}] table[x=Time,y=Track2] {data/TUD-Stadtmitte_135.txt};
  \addplot+[teal,solid,mark=*, mark size=2, line width=1pt, mark options={fill=teal}] table[x=Time,y=Track3] {data/TUD-Stadtmitte_135.txt};
\end{axis}
\end{tikzpicture}}
\vspace{-0.2cm}
\caption[caption]{\small recall@$0.7$\footnotemark of the visual interpolation vs the second keyframes selected for annotation for $4$ objects in \textit{TUD-Stadtmitte} video
 (the first keyframe is fixed, and marked as frame $0$ for simplicity). Notice that for each object a different frame should be annotated to maximize annotation quality for its track.}\label{fig:frame_selection}
\vspace{-0.4cm}
\end{figure}
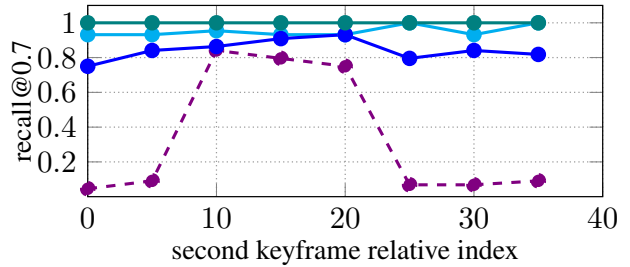
Moreover, instead of the original AlexNet backbone we use MobileNetV3~\cite{mobilenetv3} backbone (as it delivers better performance). Since MobileNetV3 is not fully convolutional we extensively use data augmentation in training, as described in~\cite{siamrpnplus}.\footnotetext{recall@$0.7$ is computed as a fraction of generated bounding boxes that have intersection-over-union(IoU) with the groundtruth higher than $0.7$}

\begin{figure*}[t]
\begin{subfigure}[c]{0.49\linewidth}
\centering
\begin{tikzpicture}
\tikzstyle{every node}=[font=\small]
 \begin{axis}[width=0.89\linewidth,height=0.5\linewidth,
              ylabel= recall@0.7,
              ylabel shift={-5pt},
              xlabel= \# manual boxes per track,
              xlabel shift={-3pt},
              ytick={0.4,0.5,...,1.1},
              xmin=2,
              xmax=10,
              ymin=0.4,
              ymax=0.9,
              enlargelimits=false,
              grid=both,
              grid style=densely dotted,
              legend pos = south east,
              legend style={nodes={scale=0.8, transform shape}},
              minor grid style={white!85!black},
              major grid style={white!60!black},
              extra tick style={white!85!black}
              ] 
  \addplot+[lightgray,dashed,mark=*, mark size=1.2, line width=1.7pt, mark options={fill=lightgray}] table[x=NumBoxPerTrack,y=Recall] {data/imnet/linea_uks_Recall07_numbox.txt};
  \addlegendentry{linear interpolation}                  
  \addplot+[cyan,dashed,mark=*, mark size=1.2, line width=1.7pt, mark options={fill=cyan}] table[x=NumBoxPerTrack,y=Recall] {data/imnet/pointnet0_uks_Recall07_numbox.txt};
  \addlegendentry{tracking}
  \addplot+[blue,dashed,mark=*, mark size=1.2, line width=1.7pt, mark options={fill=blue}] table[x=NumBoxPerTrack,y=Recall] {data/imnet/pointnet3_uks_Recall07_numbox.txt};
  \addlegendentry{visual interpolation}
\end{axis}              
\end{tikzpicture}
\end{subfigure} 
\begin{subfigure}[c]{.49\linewidth}
\centering
\begin{tikzpicture}
\tikzstyle{every node}=[font=\small]
 \begin{axis}[width=0.89\linewidth,height=0.5\linewidth,
              ylabel= recall@0.7,
              ylabel shift={-5pt},
              xlabel= \# manual boxes per track,
              xlabel shift={-3pt},
              ytick={0.4,0.5,...,1.1},
              xmin=2,
              xmax=10,
              ymin=0.4,
              ymax=0.9,
              enlargelimits=false,
              grid=both,
              grid style=densely dotted,
              legend pos = south east,
              legend style={nodes={scale=0.8, transform shape}},
              minor grid style={white!85!black},
              major grid style={white!60!black},
              extra tick style={white!85!black}
              ] 
  \addplot+[lightgray,dashed,mark=*, mark size=1.2, line width=1.7pt, mark options={fill=lightgray}] table[x=NumBoxPerTrack,y=Recall] {data/imnet/linea_uks_Recall07_numbox.txt};
  \addlegendentry{linear interpolation}                 
  \addplot+[cyan,dashed,mark=*, mark size=1.2, line width=1.7pt, mark options={fill=cyan}] table[x=NumBoxPerTrack,y=Recall] {data/imnet/rpn0_uks_Recall07_numbox.txt};
  \addlegendentry{tracking}
  \addplot+[blue,dashed,mark=*, mark size=1.2, line width=1.7pt, mark options={fill=blue}] table[x=NumBoxPerTrack,y=Recall] {data/imnet/rpn3_uks_Recall07_numbox.txt};
  \addlegendentry{visual interpolation}                
\end{axis}              
\end{tikzpicture}
\end{subfigure}
\vspace{-0.2cm}
\caption{\small Performance of the linear interpolation, tracking and visual interpolation models ($K=2$) at recall@0.7 for SiameseFC and DaSiamRPN models. Interpolation models have a clear advantage over the base tracker model.}\label{fig:tracker_ablation_modeltype}
\vspace{-0.4cm}
\end{figure*}
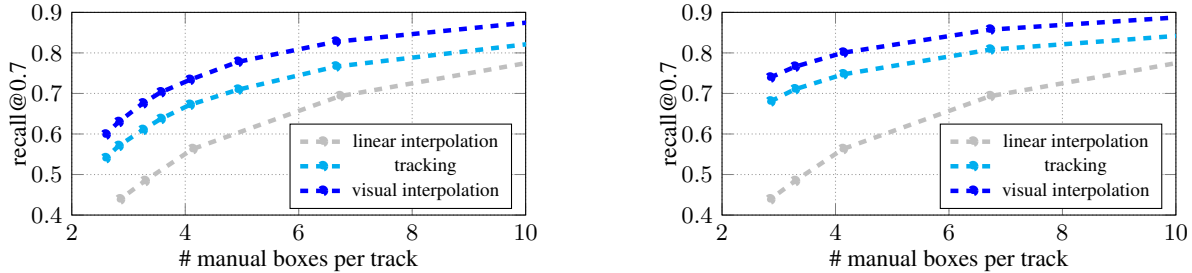

\subsection{Frame selection guidance}\label{sec:guidance_module}
As mentioned in Sec~\ref{sec:intro} and confirmed by experiments in Sec~\ref{sec:rater-experiments}, one of the major slow-downs for the annotation process is suboptimal selection of the frames to be manually annotated (\textit{keyframes}).  
In Fig~\ref{fig:frame_selection} we show that the quality of the visual interpolation model predictions clearly depends on the subset of keyframes manually annotated. To analyze this, we select a video clip containing $4$ different objects and investigate the quality of annotation for each object depending on the selected second keyframe (the first keyframe is the same for all objects). For each object the optimal second keyframe is different and it has large impact on the annotation quality (depending on the object, quality increases by up to $+70\%$ when selecting the optimal keyframe, compared to the worst keyframe).
We propose here to optimize the annotation process by introducing an automatic frame selection mechanism.
Given already existing annotations of an object in some previous frames, we want to select the {\em next keyframe} that would maximize the quality of the annotations produced by our visual interpolation module in the unannotated portion of the video.
In this way we avoid the need to jump back and forth across the timeline, which can confuse the annotator and requires expensive context switching~\cite{pathtrack,vatical}.
In~\cite{bubblenets} the authors proposed an architecture to select a single best frame to propagate a segmentation mask to the whole video sequence. However, their approach operates on the full frames and therefore lacks an important element --- conditioning on a specific target object. We extend their approach by introducing an attention mechanism to condition the model predictions on the object to be annotated.
\mypar{Method overview.}
Our method works as follows.
First, we sample candidate keyframes uniformly in an interval of 100 frames after all previously annotated frames.
Then, we rank these candidate keyframes by expected annotation quality.
At the core of our approach we train a ranking model that operates on pairs of candidate keyframes. It predicts a score indicating which of the two candidates is better, conditioned on the appearance of a specific target object, as captured by bounding boxes in previously annotated frames. The ranking model also takes into account the unannotated video content.
The final score for each candidate keyframe is calculated as the sum over all pairwise scores. The single top-scoring candidate is selected as the next keyframe. The annotator then manually draws the object bounding box on this keyframe, and the process iterates.

\mypar{Ranking model architecture.}
Fig~\ref{fig:guidance} illustrates the architecture of our model. 
It takes three kinds of input:
(1) a pair of candidate keyframes;
(2) a set of $N$ {\em reference frames} randomly sampled from the unannotated part of the video, enabling to condition on the content of the video; and
(3) $K-1$ frames cropped around the bounding box from previously annotated frames ({\em templates}), enabling to condition on previous annotations for this object.
We use a a fully convolutional feature extractor to extract features from the full candidate and reference frames ($\{f^j\}_{j=1}^{N+2}$) and the templates ($\{z^j\}_{j=1}^{K-1}$). We implement conditioning on templates by computing attention maps $a_j$.
These are computed by cross-correlation ($*$) between template features and the respective video frame features ($g(\cdot)$ denotes max-pooling):
\begin{align*}
 a_j = g(\varphi(z_1),\cdots, &\varphi(z_{K-1})) * \varphi(f^j)
\end{align*}
The attention maps help to ensure that the module is focusing on the relevant parts of the image (i.e. on the target object, whose appearance is captured by the template features).
The final prediction for a pair of candidate keyframes is a single score computed by several fully convolutional layers ($F'(\cdot)$) operating on top of the extracted features and attention maps (the scores are normalized to $[-1,1]$):
\begin{align*}
 c =  F'([a_1 + \varphi'(f_1),\dots, a_{N+2} + \varphi'(f_{N+2}) ])
\end{align*}

\mypar{Quality score for a candidate keyframe.}
We run the ranking model for all pairs of candidate keyframes.
The overall score of a candidate keyframe is computed as the sum of all positive comparison scores (i.e. for pairs where this candidate keyframe was better than the frame it was compared against).
The candidate keyframes are then sorted by their overall scores and the highest-scoring one is selected as the next keyframe to be annotated.
%

Although the proposed approach is related to~\cite{bubblenets}, it goes well beyond. Thanks to the newly introduced conditioning on the target object we are able to handle the more complex (and realistic) scenario where the prediction must be done not simply at the frame level but for a specific object (see Fig.~\ref{fig:frame_selection}).
In Sec.~\ref{sec:simulations} we show that conditioning is crucial for the performance of the ranking model.

\mypar{Training.}
The ranking model is trained in a supervised manner. To obtain training labels, we:
(1) randomly sample previously annotated frames (templates) and pairs of candidate keyframes;
(2) run the visual interpolation model for each candidate keyframe in a pair, and then evaluate its predictions over a $100$ frame interval against ground-truth bounding boxes.
The difference between the visual interpolation predictions quality (recall@$0.7$) of the two candidates is used as binary label for training the ranking model.
To reduce noise in the training data, we only consider tracks of objects larger than $5\%$ of the frame area.
Moreover, for a given template we sample multiple pairs of candidate keyframes such that there is a significant difference in the quality of the visual interpolation predictions they lead to (empirically set to $>0.3$]).

The model is trained with binary cross entropy loss. We employ a feature extractor similar to AlexNet~\cite{alexnet}, described in~\cite{siamesefc}.
The ranking model is trained for $10$ epochs using momentum optimizer~\cite{momentum} with $1e-3$ initial leaning rate and batch size $12$. In general we observed better training stability with larger batch size, which confirms findings by~\cite{noisylabels} that larger batch sizes improve training on noisy labels.

\section{Experimental results}

First, we evaluate the performance of our framework on the ImageNet VID validation set~\cite{imagenetvid} (Sec.~\ref{sec:simulations}).
Second, we evaluate the proposed framework by running annotation process with human annotators on Got10k validation set~\cite{got10k} (Sec.~\ref{sec:rater-experiments}) and analysing results of human annotator experiments vs. simulation predictions.
Finally, we compare  the proposed method with state-of-the-art approaches~\cite{pathtrack,labelme2016,vatic,vatical} on MOT2015 dataset~\cite{mot15} and demonstrate generalization across datasets (Sec.~\ref{sec:other-tools}).
\begin{figure}
\centering
\includegraphics[width=0.45\linewidth]{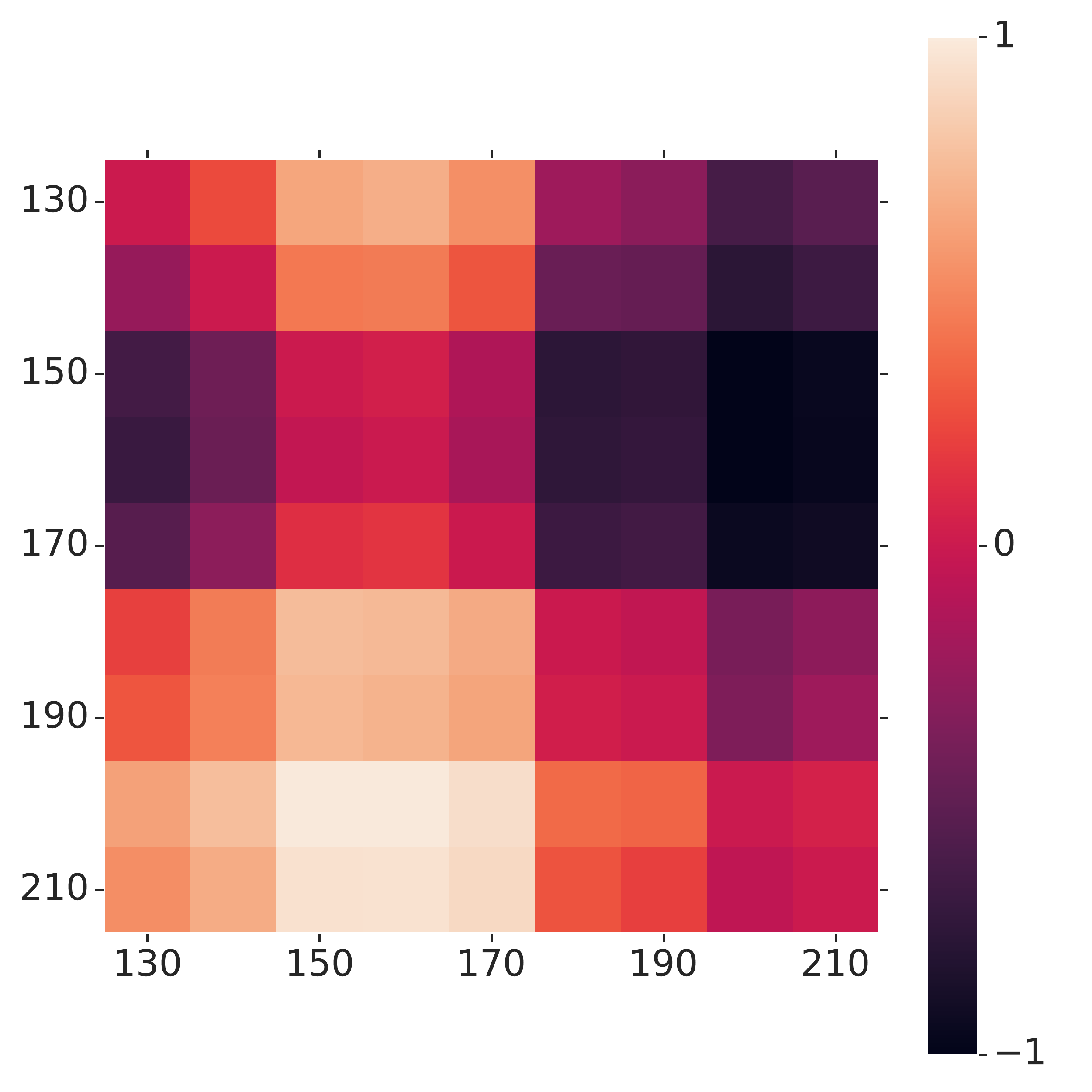}
\includegraphics[width=0.45\linewidth]{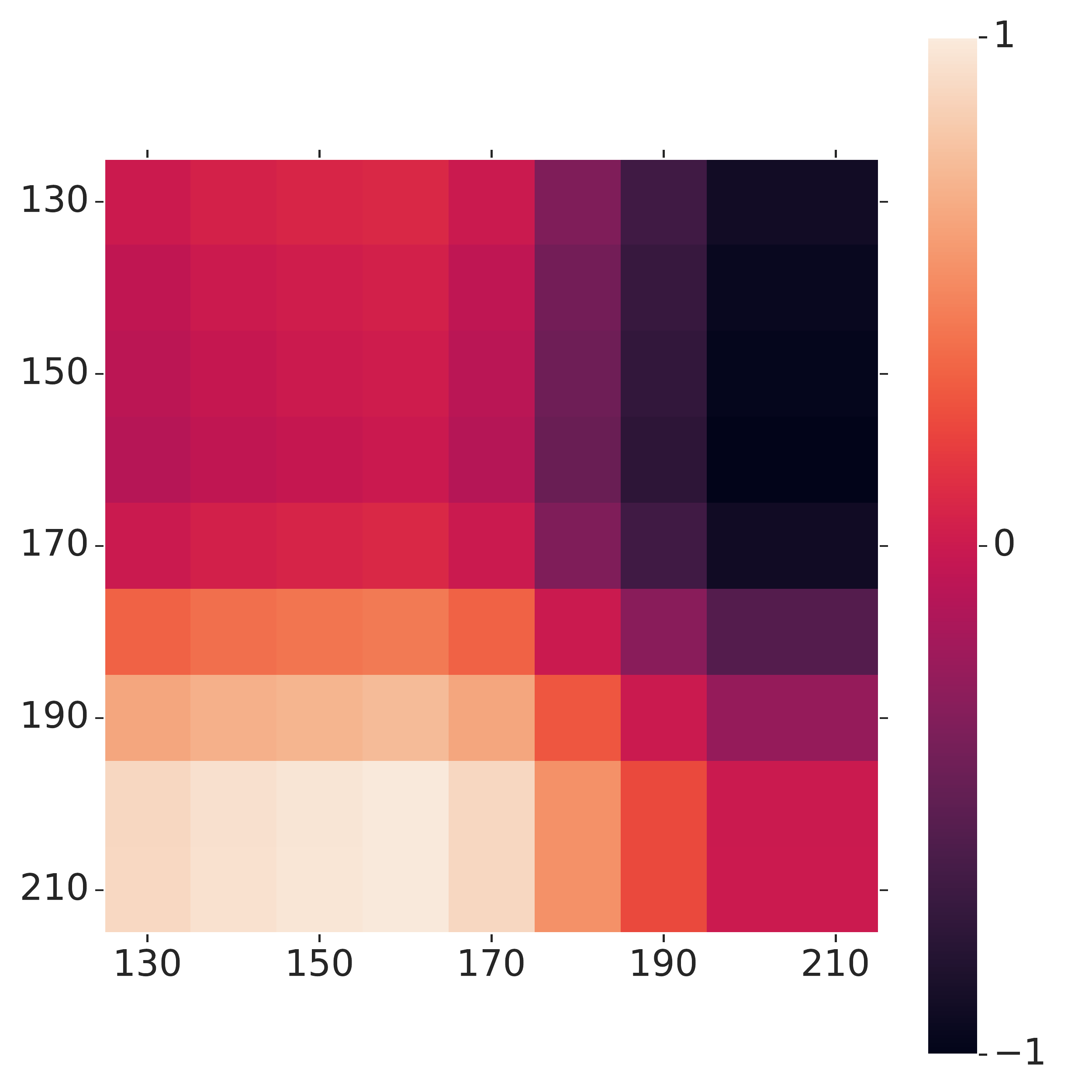}
\vspace{-0.2cm}
\caption{\small Frame comparison matrix --- vertical and horizontal axis represent frame offsets from the already annotated frames; values at cell $i,j$ represent relative annotation quality after annotating frame $i$ or frame $j$: left --- model prediction; right --- groundtruth.}\label{fig:modelvsgt}
\vspace{-0.4cm}
\end{figure}
\subsection{Performance of the framework components}
\label{sec:simulations}
ImageNet VID~\cite{imagenetvid} is a middle-scale video object tracking dataset with dense trajectory annotations. 
The training set contains $3862$ videos and objects of $30$ classes. On average, each video contains $2.35$ object tracks (with maximum of $47$) and the average object size is $16\%$ of the image area.
We evaluate on the validation set, which contains $555$ videos.

\mypar{Results for visual interpolation.}
We show that our proposed extension of the tracker models (Sec.~\ref{sec:visual-interpolation} is applicable to several contemporary deep tracker architectures and consistently increases model performance compared to tracking).
We train all configurations of the model with $K=2$.
We compare visual interpolation to linear interpolation and a forward tracking model as widely used baselines. As a metric, we  plot the recall@0.7 curve as a function of  the average number of manual boxes annotated per object track.
For this comparison we uniformly sample keyframes at different sampling intervals.
Fig~\ref{fig:tracker_ablation_modeltype} shows that visual interpolation works clearly better than linear interpolation and tracking. 
\begin{table}
\centering
\begin{tabular}{l cc}
\toprule
Model name & all &  no small obj \\
\midrule
no attention & $0.51$ & $0.51$\\
no vis. features & $0.56$& $0.61$\\
full model & $\textbf{0.63}$ & $\textbf{0.68}$\\
\bottomrule
\end{tabular}
\caption{\small Ranking model accuracy: model with \textit{no attention} uses visual features only; \textit{no vis. features} model only uses attention maps; \textit{full model} is the full model as in Section~\ref{sec:guidance_module}; the {\em no small obj} column reports accuracy for objects with area $>15\%$ of the image.}\label{tab:fg_acc}
\vspace{-0.4cm}
\end{table}
We choose DaSiamRPN visual interpolation as the model with better performance for further experiments.

\mypar{Results for frame selection guidance.}
First, to motivate the choice of model architecture, we compare the performance of three variations: the architecture without attention, the architecture without visual features, and  the full model. We compare them in terms of binary classification accuracy.
More precisely, we randomly sample pairs of test frames from the validation set, such that
(1) the difference in performance between two frames within a pair is significant, and (2) the number of pairs where the first frame performs better than second is balanced (i.e. a random classifier produces accuracy $0.5$).
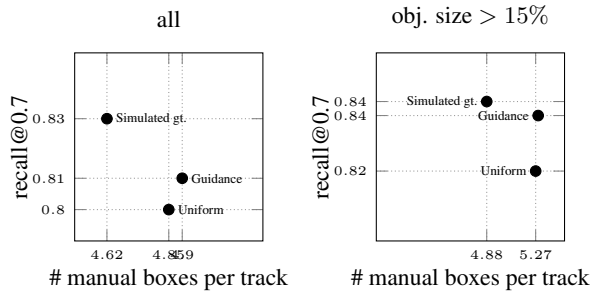
\begin{figure}
\begin{tikzpicture}
\tikzstyle{every node}=[font=\tiny]
\begin{axis}[width=0.49\linewidth,height=0.49\linewidth,
              ylabel=recall@0.7,
              ylabel shift={-5pt},
              xlabel=\# manual boxes per track,
              xlabel shift={-3pt},
              label style={font=\small},
              ytick={0.800, 0.81, 0.829},
              xmin = 4.5,
              xmax = 5.2,
              ymin = 0.79,
              ymax = 0.85,
              xtick={4.62,  4.90, 4.85},
              enlargelimits=false,
              title = {all},
              title style = {font=\small},
              grid=both,
              grid style=densely dotted,
              legend pos = south east,
              legend style={nodes={scale=0.8, transform shape}},
              minor grid style={white!85!black},
              major grid style={white!60!black},
              extra tick style={white!85!black}
              ]
  \filldraw[black] (4.62, 0.829) circle (2pt) node[anchor=west] {Simulated gt.};
  \filldraw[black] (4.90, 0.81) circle (2pt) node[anchor=west] {Guidance};
  \filldraw[black] (4.85, 0.800) circle (2pt) node[anchor=west] {Uniform}; 
\end{axis}
\end{tikzpicture}
\begin{tikzpicture}
\tikzstyle{every node}=[font=\tiny]
\begin{axis}[width=0.49\linewidth,height=0.49\linewidth,
              ylabel=recall@0.7,
              ylabel shift={-5pt},
              xlabel=\# manual boxes per track,
              xlabel shift={-3pt},
              label style={font=\small},
              ytick={0.8222937402, 0.8399995887, 0.8444723205},
              xmin = 4,
              xmax = 5.5,
              ymin = 0.80,
              ymax = 0.86,
              xtick={4.88, 5.27},
              title = {obj. size $>15\%$},
              title style = {font=\small},
              enlargelimits=false,
              grid=both,
              grid style=densely dotted,
              legend pos = south east,
              legend style={nodes={scale=0.8, transform shape}},
              minor grid style={white!85!black},
              major grid style={white!60!black},
              extra tick style={white!85!black}
              ]
  \filldraw[black] (4.88, 0.8444723205) circle (2pt) node[anchor=east] {Simulated gt.};
  \filldraw[black] (5.29, 0.8399995887) circle (2pt) node[anchor=east] {Guidance};
  \filldraw[black] (5.27, 0.8222937402) circle (2pt) node[anchor=east] {Uniform}; 
\end{axis}
\end{tikzpicture}
\vspace{-0.2cm}
\caption{\small Frame selection guidance vs uniform sampling (in terms of recall@0.7);
{\em simulated gt.}: ground truth is used for frame selection guidance;
{\em guidance}: visual interpolation with keyframes predicted by our guidance module;
{\em uniform}: visual interpolation with uniform keyframe selection.}\label{fig:ablation_imagenet}
\end{figure}

The results are presented in Table~\ref{tab:fg_acc}. Our full model clearly wins against both baseline models. Further, the model using no attention does not do better than random chance. The larger gap for the test sample that does not contain small objects is probably explained by the fact that the smaller is an object, the more noisy are the labels on the validation set.
Fig.~\ref{fig:modelvsgt} shows the pairwise comparison matrix predicted by the model and the ground truth matrix that evaluates which frames are better to manually annotate so that the visual interpolation model would work better. Interestingly, the model confidence in the frame comparison correlates with the performance difference in the ground-truth, although the model is trained for classification.
We further show the improvement from using the frame selection guidance module in the full experiment (Fig~\ref{fig:ablation_imagenet}).
We compare running the visual interpolation module using uniformly spaced keyframes, versus with frame selection guidance.
We also show guidance based on ground-truth signal for comparison (albeit it 
 does not imply globally optimal keyframe selection per track). 
As can be seen, our frame selection module outperforms the uniformly sampling frames and delivers bigger improvement for the subset that does not contain small objects.
Overall, we point out that the problem of predicting model performance is a very challenging task, hence even $~2\%$ improvement is significant and can result in hours of annotation time spared.
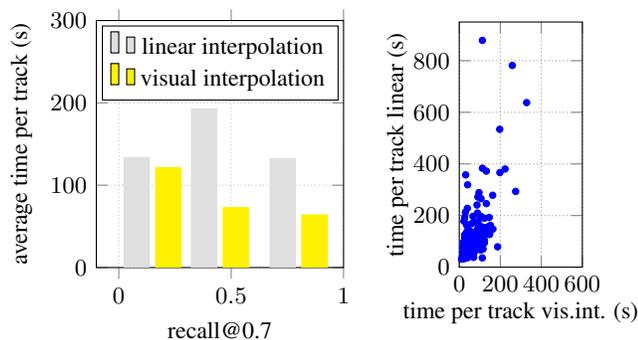
\begin{figure}[t]
\begin{subfigure}[c]{.59\linewidth}
\begin{tikzpicture}
\tikzstyle{every node}=[font=\small]
 \begin{axis}[ybar,width=0.99\linewidth,height=0.99\linewidth,
              ylabel=average time per track (s),
              xlabel=recall@0.7,
              xmin=-0.1,
              xmax=1.0,
              ymin=0,
              ymax=300,
              enlargelimits=false,grid=both,grid style=densely dotted]
  \addplot [ybar, bar width=10pt,draw=none, fill=lightgray!50] table[x=Bin,y=LinearTime,col sep=tab] {data/ratings/siamese_vs_linear_binned.txt};
  \addlegendentry{linear interpolation}
  \addplot [ybar, bar width=10pt,draw=none, fill=yellow!100] table[x=Bin,y=SiameseTime,col sep=tab] {data/ratings/siamese_vs_linear_binned.txt};
  \addlegendentry{visual interpolation}
  \end{axis}
\end{tikzpicture}
\end{subfigure} 
\begin{subfigure}[c]{.39\linewidth}
\begin{center}
\begin{tikzpicture}
\tikzstyle{every node}=[font=\small]
\begin{axis}[width=0.99\linewidth,height=1.49\linewidth,
              ylabel= time per track linear (s),
              ylabel shift={-5pt},
              xlabel= time per track vis.int. (s),
              xlabel shift={-3pt},
              label style={font=\small}, 
              title style = {font=\small},
              ymin=0.0,
              ymax=950.0, 
              xmin=0.0,
              xmax=600.0,                             
              enlargelimits=false,
              grid=both,
              grid style=densely dotted,
              legend pos = south east,
              legend style={nodes={scale=0.8, transform shape}},
              minor grid style={white!85!black},
              major grid style={white!60!black},
              extra tick style={white!85!black}
              ]
  \addplot[blue, only marks,mark=*, mark size=1.2, mark options={fill=blue}] table[x=Siamese,y=Linear] {data/ratings/siamese_vs_linear.txt};              
\end{axis}
\end{tikzpicture}
\end{center}
\end{subfigure}
\vspace{-0.2cm}
\caption{\small Left: linear interpolation vs visual interpolation: annotation time at three quality levels (lower is better). Right: annotation time comparison between the two approaches, where each blue dot is a different video.}\label{fig:linear_vs_interpolation}
\vspace{-0.4cm}
\end{figure}

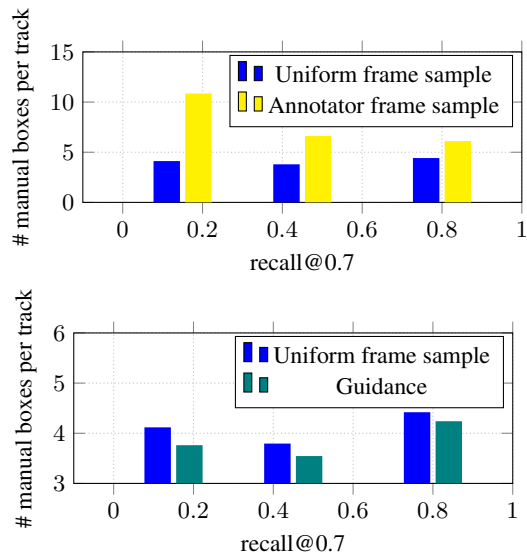
\begin{figure}
\center
\small
\begin{tikzpicture}
\tikzstyle{every node}=[font=\small]
 \begin{axis}[ybar,width=0.89\linewidth,height=0.43\linewidth,
              ylabel=\# manual boxes per track,
              xlabel=recall@0.7,
              xmin=-0.1,
              xmax=1.0,
              ymin=0,
              ymax=15,
              enlargelimits=false,
              grid=both,
              grid style=densely dotted]
  \addplot [ybar, bar width=10pt,draw=none, fill=blue] table[x=Bin,y=SiameseSim,col sep=tab] {data/ratings/siamese_sim_binned.txt};
  \addlegendentry{Uniform frame sample};
  \addplot [ybar, bar width=10pt,draw=none, fill=yellow!100] table[x=Bin,y=Siamese,col sep=tab] {data/ratings/siamese_sim_binned.txt};
  \addlegendentry{Annotator frame sample};
  \end{axis}
\end{tikzpicture}
\begin{tikzpicture}
\tikzstyle{every node}=[font=\small]
 \begin{axis}[ybar,width=0.89\linewidth,height=0.43\linewidth,
              ylabel=\# manual boxes per track,
              xlabel=recall@0.7,
              xmin=-0.1,
              xmax=1.0,
              ymin=3,
              ymax=6,
              enlargelimits=false,
              grid=both,
              grid style=densely dotted]
  \addplot [ybar, bar width=10pt,draw=none, fill=blue] table[x=Bin,y=SiameseSim,col sep=tab] {data/ratings/siamese_different_simulations.txt};
  \addlegendentry{Uniform frame sample};
  \addplot [ybar, bar width=10pt,draw=none, fill=teal!100] table[x=Bin,y=SiameseOptPr,col sep=tab] {data/ratings/siamese_different_simulations.txt};  
  \addlegendentry{Guidance};  
  \end{axis}
\end{tikzpicture}
\vspace{-0.2cm}
\caption{\small Top: number of manually drawn boxes per object track vs recall@0.7 for simulated annotations at uniform frame sampling (every $40$th frame) vs annotations by human annotators. Bottom: simulated annotation at uniform frame sampling (every $40$th frame) vs annotation with frame selection guidance.}\label{fig:human_vs_optimal}
\vspace{-0.4cm}
\end{figure}
\subsection{Experiments with human annotators}
\label{sec:rater-experiments}

Simulations do not provide full insights into the actual benefits and drawbacks of the proposed approach when used in practice. Hence we set up a video annotation experiment with human annotators.
We use the validation set of the Got10k~\cite{got10k} dataset and compare the results obtained by annotators with the simulation results. Got10k is a highly diverse dataset containing in total $563$ classes, hence we are able to demonstrate the generalization properties of our model.
Got10k validation set contains $180$ videos, with a single annotated object in each video. 
We perform human studies with $10$ human annotators. Each annotator is asked to annotate the same set of videos with two annotation methods. The target object is defined by a bounding box annotation in the first frame of each video. 
The annotators are given a quality target of $70\%$ overlap with (hidden ideal) groundtruth box in each frame and recommended time per question of $2$ minutes.
Fig~\ref{fig:linear_vs_interpolation} presents the results of the linear vs visual interpolation comparison. With visual interpolation the annotators are able to achieve significant speedup at all quality level considered. Moreover, overall across all annotators and videos in the dataset, visual interpolation reduced annotation cost by about $50\%$: it took total of $~6.96$ hours to annotate the dataset with linear interpolation and only $3.45$ hours with visual interpolation. The average quality of annotations in terms of recall@0.7 is $0.73$ for linear interpolation and $0.75$ for visual interpolation.
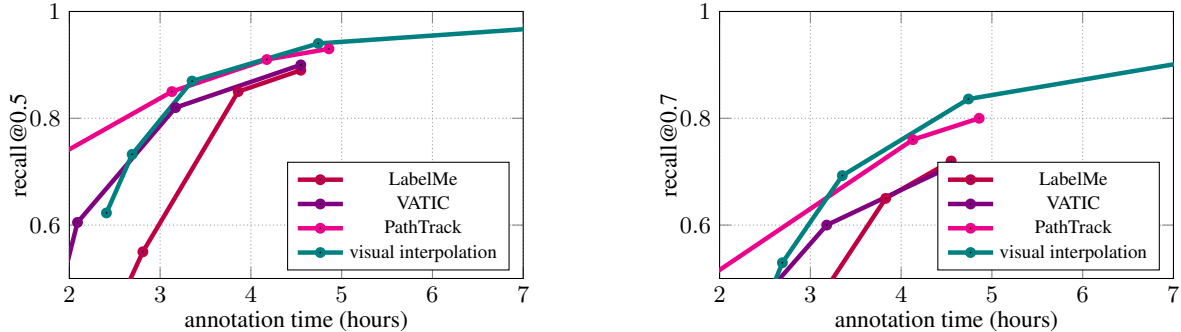
\begin{figure*}[]
\begin{subfigure}[c]{.49\linewidth}
\small
\begin{tikzpicture}
\tikzstyle{every node}=[font=\small]
\begin{axis}[width=0.89\linewidth,height=0.6\linewidth,
              ylabel=recall@0.5,
              ylabel shift={-5pt},
              xlabel=annotation time (hours),
              xlabel shift={-3pt},
              label style={font=\small}, 
              title style = {font=\small},              
              ytick={0.2,0.4,...,1.0},
              xmin=2,
              xmax=7,
              ymin=0.5,
              ymax=1.0,
              enlargelimits=false,
              grid=both,
              grid style=densely dotted,
              legend pos = south east,
              legend style={nodes={scale=0.8, transform shape}},
              minor grid style={white!85!black},
              major grid style={white!60!black},
              extra tick style={white!85!black}
              ]   
  \addplot+[purple,solid,mark=*, mark size=1.2, line width=1.7pt, mark options={fill=blue}] table[x=NumBoxPerTrack,y=Recall] {data/recall_vs_time_0.5/labelme.txt};
  \addlegendentry{LabelMe}
  \addplot+[violet,solid,mark=*, mark size=1.2, line width=1.7pt, mark options={fill=black}] table[x=NumBoxPerTrack,y=Recall] {data/recall_vs_time_0.5/vatic.txt};
  \addlegendentry{VATIC}              
  \addplot+[magenta,solid,mark=*, mark size=1.2, line width=1.7pt, mark options={fill=black}] table[x=NumBoxPerTrack,y=Recall] {data/recall_vs_time_0.5/pathtrack.txt};
  \addlegendentry{PathTrack} 
  \addplot+[teal,solid,mark=*, mark size=1.2, line width=1.7pt, mark options={fill=black}] table[x=Time,y=Recall] {data/rpn3_Recall05_time.txt};
  \addlegendentry{visual interpolation}  
\end{axis}
\end{tikzpicture}
\end{subfigure}
\begin{subfigure}[c]{.49\linewidth}
\small
\begin{tikzpicture}
\tikzstyle{every node}=[font=\small]
\begin{axis}[width=0.89\linewidth,height=0.6\linewidth,
              ylabel=recall@0.7,
              ylabel shift={-5pt},
              xlabel=annotation time (hours),
              xlabel shift={-3pt},
              label style={font=\small}, 
              title style = {font=\small},              
              ytick={0.2,0.4,...,1.0},
              xmin=2,
              xmax=7,
              ymin=0.5,
              ymax=1.0,
              enlargelimits=false,
              grid=both,
              grid style=densely dotted,
              legend pos = south east,
              legend style={nodes={scale=0.8, transform shape}},
              minor grid style={white!85!black},
              major grid style={white!60!black},
              extra tick style={white!85!black}
              ]   
  \addplot+[purple,solid,mark=*, mark size=1.2, line width=1.7pt, mark options={fill=blue}] table[x=NumBoxPerTrack,y=Recall] {data/recall_vs_time_0.7/labelme.txt};
  \addlegendentry{LabelMe}
  \addplot+[violet,solid,mark=*, mark size=1.2, line width=1.7pt, mark options={fill=black}] table[x=NumBoxPerTrack,y=Recall] {data/recall_vs_time_0.7/vatic.txt};
  \addlegendentry{VATIC}              
  \addplot+[magenta,solid,mark=*, mark size=1.2, line width=1.7pt, mark options={fill=black}] table[x=NumBoxPerTrack,y=Recall] {data/recall_vs_time_0.7/pathtrack.txt};
  \addlegendentry{PathTrack}     
  \addplot+[teal,solid,mark=*, mark size=1.2, line width=1.7pt, mark options={fill=black}] table[x=Time,y=Recall] {data/rpn3_Recall07_time.txt};
  \addlegendentry{visual interpolation}     
\end{axis}
\end{tikzpicture}
\end{subfigure}
\vspace{-0.2cm}
\caption{\small Annotation time required to annotate MOT2015 dataset with a given quality in terms of recall@0.5 and recall@0.7.}\label{fig:performanceMOTTime}
\vspace{-0.4cm}
\end{figure*}
The annotations were not given any specific guidelines as to how to select which frames should be annotated manually. For both visual and linear interpolation they relied on their understanding of which frames should be annotated. 
Next, we investigate how well the annotators select which frame to annotate.
In Fig~\ref{fig:human_vs_optimal} we compare the selection made by human annotators vs uniform frame sampling. 
We can clearly see that humans lack the ability to select frames optimally: even uniform frame selection with constant sampling interval ($40$ in this experiment) leads to a faster annotation process, or better annotation quality at the same speed.
Finally, we also evaluate efficiency gains from applying our frame selection guidance mechanism (Sec.~\ref{sec:guidance_module}).
In Fig~\ref{fig:human_vs_optimal}-right we show that frame selection model allows to surpass the performance of uniform sampling and improves over the baseline where humans select the frames to annotate themselves.  These show that frame selection model delivers on average $~6.5\%$ reduction in the number of manual boxes needed, at no loss in quality.
Those results demonstrate the importance of the good models for frame selection for the annotation process, as for large-scale annotation even small improvement can bring significant cost savings.

\subsection{Comparison to other annotation tools}
\label{sec:other-tools}
In this section, we compare our full method to other annotation tools~\cite{pathtrack,vatic,labelme2016} on the MOT2015~\cite{mot15} dataset.
The training set contains $11$ video sequences with an average of $45$ tracks per video. The dataset contains only annotations for the class "person" but some videos contain $100+$ annotated tracks, creating challenging setting for single-object tracking algorithms.
We compare to the results reported in~\cite{pathtrack} (for PathTrack, as well as for VATIC~\cite{vatic} and LabelMe~\cite{labelme2016}), as they performed a comprehensive evaluation of their approach and compare to several other state-of-the-art annotation tools.
To perform the comparison, we estimate the actual annotation time based on the time measurements provided in~\cite{pathtrack} and the number of boxes drawn manually in our protocol. According to~\cite{pathtrack}, the average time to draw a box is $t_{box} = 5.2$s and the total annotation time is calculated as:
\begin{equation}
  t_{track} = \lambda t_{watch} +  t_{box} \cdot N_{box} 
\end{equation}
where $t_{watch}$ is the time for watching through a track, $t_{track}$ is the annotation time per track and $N_{box}$ is the number of boxes the annotator has drawn. 
The results are presented in Fig.~\ref{fig:performanceMOTTime} on two metrics: recall@0.5 and recall@0.7 versus annotation time.
Fig~\ref{fig:performanceMOTTime} shows that, when collecting many boxes of high quality our method outperforms all provided baselines. For example, at $80\%$ of the data annotated with quality of $0.7$ IoU or higher, we achieve a $10\%$ reduction of the annotation time compared to the strongest baseline (PathTrack). The more the required annotation quality increases, the bigger is the advantage of our method in terms of annotation time. 
We want to underline that PathTrack~\cite{pathtrack} is designed as a method for fast but imprecise annotation, while our method is designed for obtaining more accurate annotations and hence each method serves a different purpose. Further, our method is generic (not specific to the 'person' class) and does not require post-processing of the data (PathTrack needs to align automatically detected boxes with annotated object tracks). 
For example, compared to VATIC~\cite{vatic} and LabelMe~\cite{labelme2016}, we achieve $33\%$ speedup for the fixed quality of $70\%$ of the boxes annotated with quality of $0.7$ IoU or higher. 

\section{Conclusions}
We presented and evaluated a unified framework for interactive video bounding box annotation. We introduced a visual interpolation algorithm which is based on contemporary trackers but allows for track correction. Moreover, we  presented a frame selection guidance module and experimentally showed its importance within the annotation process.
We evaluated (in simulations) that using a visual signal allows to annotate $60\%$ less boxes than the traditionally used linear interpolation while keeping the same quality.
In experiments with human annotators we have shown that annotation time can be reduced by more than $50\%$ using the proposed framework. Further, we also showed that proposed approach saves $10\%$ of annotation time compared to the state-of-the-art method Pathtrack (and more compared to LabelMe~\cite{labelme2016} and VATIC~\cite{vatic}) on challenging multi-object tracking dataset MOT2015~\cite{mot15}.

{\small
\bibliographystyle{ieee_fullname}
\bibliography{egbib}

\begin{thebibliography}{10}\itemsep=-1pt

\bibitem{siamesefc}
Luca Bertinetto, Jack Valmadre, Jo{\~a}o~F Henriques, Andrea Vedaldi, and
  Philip~HS Torr.
\newblock Fully-convolutional siamese networks for object tracking.
\newblock {\em arXiv preprint arXiv:1606.09549}, 2016.

\bibitem{DimP}
Goutam Bhat, Martin Danelljan, Luc~Van Gool, and Radu Timofte.
\newblock Learning discriminative model prediction for tracking.
\newblock {\em CoRR}, abs/1904.07220, 2019.

\bibitem{scribblebox}
Bowen Chen, Huan Ling, Xiaohui Zeng, Gao Jun, Ziyue Xu, and Sanja Fidler.
\newblock Scribblebox: Interactive annotation framework for video object
  segmentation.
\newblock In {\em European Conference on Computer Vision (ECCV)}, 2020.

\bibitem{meanshift}
Dorin Comaniciu and Peter Meer.
\newblock Mean shift: A robust approach toward feature space analysis.
\newblock {\em IEEE Trans. Pattern Anal. Mach. Intell.}, 24(5):603--619, May
  2002.

\bibitem{lasot}
Heng Fan, Liting Lin, Fan Yang, Peng Chu, Ge Deng, Sijia Yu, Hexin Bai, Yong
  Xu, Chunyuan Liao, and Haibin Ling.
\newblock Lasot: A high-quality benchmark for large-scale single object
  tracking.
\newblock In {\em The IEEE Conference on Computer Vision and Pattern
  Recognition (CVPR)}, June 2019.

\bibitem{geomboxint}
Pedro Gil-Jim{\'e}nez, Hilario G{\'o}mez-Moreno, Roberto L{\'o}pez-Sastre, and
  Saturnino Maldonado-Basc{\'o}n.
\newblock Geometric bounding box interpolation: an alternative for efficient
  video annotation.
\newblock {\em EURASIP Journal on Image and Video Processing}, 2016.

\bibitem{bubblenets}
Brent~A. Griffin and Jason~J. Corso.
\newblock Bubblenets: Learning to select the guidance frame in video object
  segmentation by deep sorting frames.
\newblock In {\em The IEEE Conference on Computer Vision and Pattern
  Recognition (CVPR)}, June 2019.

\bibitem{ava}
Chunhui Gu, Chen Sun, David Ross, Carl Vondrick, Caroline Pantofaru, Yeqing Li,
  Sudheendra Vijayanarasimhan, George Toderici, Susanna Ricco, Rahul
  Sukthankar, Cordelia Schmid, and Jitendra Malik.
\newblock Ava: A video dataset of spatio-temporally localized atomic visual
  actions.
\newblock pages 6047--6056, 06 2018.

\bibitem{lvis}
Agrim Gupta, Piotr Dollar, and Ross Girshick.
\newblock {LVIS}: A dataset for large vocabulary instance segmentation.
\newblock In {\em Proceedings of the {IEEE} Conference on Computer Vision and
  Pattern Recognition}, 2019.

\bibitem{KCFs}
Joao Henriques, Rui Caseiro, Pedro Martins, and Jorge Batista.
\newblock High-speed tracking with kernelized correlation filters.
\newblock {\em IEEE Transactions on Pattern Analysis and Machine Intelligence},
  37, 04 2014.

\bibitem{CFWCR}
Junfei Zhuang Yuan Dong Hongliang~Bai hiqun He, Yingruo~Fan.
\newblock Correlation filters with weighted convolution responses.
\newblock {\em IEEE International Conference on Computer Vision}, 2017.

\bibitem{mobilenetv3}
Andrew Howard, Mark Sandler, Grace Chu, Liang-Chieh Chen, Bo Chen, Mingxing
  Tan, Weijun Wang, Yukun Zhu, Ruoming Pang, Vijay Vasudevan, Quoc~V. Le, and
  Hartwig Adam.
\newblock Searching for mobilenetv3, 2019.

\bibitem{got10k}
Lianghua Huang, Xin Zhao, and Kaiqi Huang.
\newblock Got-10k: {A} large high-diversity benchmark for generic object
  tracking in the wild.
\newblock {\em CoRR}, 2018.

\bibitem{CONDENSATION}
Michael Isard and Andrew Blake.
\newblock Condensation -- conditional density propagation for visual tracking.
\newblock {\em INTERNATIONAL JOURNAL OF COMPUTER VISION}, 29:5--28, 1998.

\bibitem{supervoxelprop}
Suyog Jain and Kristen Grauman.
\newblock Supervoxel-consistent foreground propagation in video.
\newblock 2014.

\bibitem{Kristan2019a}
Matej Kristan, Jiri Matas, Ales Leonardis, Michael Felsberg, Roman Pflugfelder,
  Joni-Kristian Kamarainen, Luka \v{C}ehovin Zajc, Ondrej Drbohlav, Alan
  Lukezic, Amanda Berg, Abdelrahman Eldesokey, Jani Kapyla, and Gustavo
  Fernandez.
\newblock The seventh visual object tracking vot2019 challenge results, 2019.

\bibitem{alexnet}
Alex Krizhevsky, Ilya Sutskever, and Geoffrey~E Hinton.
\newblock Imagenet classification with deep convolutional neural networks.
\newblock In F. Pereira, C.~J.~C. Burges, L. Bottou, and K.~Q. Weinberger,
  editors, {\em Advances in Neural Information Processing Systems 25}, pages
  1097--1105. Curran Associates, Inc., 2012.

\bibitem{eodh}
Alina Kuznetsova, Sung Ju~Hwang, Bodo Rosenhahn, and Leonid Sigal.
\newblock Expanding object detector's horizon: Incremental learning framework
  for object detection in videos.
\newblock In {\em The IEEE Conference on Computer Vision and Pattern
  Recognition (CVPR)}, June 2015.

\bibitem{openimagesv4}
Alina Kuznetsova, Hassan Rom, Neil Alldrin, Jasper R.~R. Uijlings, Ivan Krasin,
  Jordi Pont{-}Tuset, Shahab Kamali, Stefan Popov, Matteo Malloci, Tom Duerig,
  and Vittorio Ferrari.
\newblock The open images dataset {V4:} unified image classification, object
  detection, and visual relationship detection at scale.
\newblock {\em CoRR}, 2018.

\bibitem{mot15}
Laura Leal{-}Taix{\'{e}}, Anton Milan, Ian~D. Reid, Stefan Roth, and Konrad
  Schindler.
\newblock Motchallenge 2015: Towards a benchmark for multi-target tracking.
\newblock {\em CoRR}.

\bibitem{siamrpnplus}
Bo Li, Wei Wu, Qiang Wang, Fangyi Zhang, Junliang Xing, and Junjie Yan.
\newblock Siamrpn++: Evolution of siamese visual tracking with very deep
  networks.
\newblock {\em arXiv preprint arXiv:1812.11703}, 2018.

\bibitem{siameserpn}
Bo Li, Junjie Yan, Wei Wu, Zheng Zhu, and Xiaolin Hu.
\newblock High performance visual tracking with siamese region proposal
  network.
\newblock In {\em The IEEE Conference on Computer Vision and Pattern
  Recognition (CVPR)}, June 2018.

\bibitem{siamrpn}
B. {Li}, J. {Yan}, W. {Wu}, Z. {Zhu}, and X. {Hu}.
\newblock High performance visual tracking with siamese region proposal
  network.
\newblock In {\em 2018 IEEE/CVF Conference on Computer Vision and Pattern
  Recognition}, 2018.

\bibitem{mscoco}
Tsung-Yi Lin, Michael Maire, Serge Belongie, James Hays, Pietro Perona, Deva
  Ramanan, Piotr Doll{\'a}r, and C.~Lawrence Zitnick.
\newblock Microsoft coco: Common objects in context.
\newblock In David Fleet, Tomas Pajdla, Bernt Schiele, and Tinne Tuytelaars,
  editors, {\em Computer Vision -- ECCV 2014}, pages 740--755, Cham, 2014.
  Springer International Publishing.

\bibitem{pathtrack}
Santiago Manen, Michael Gygli, Dengxin Dai, and Luc {Van Gool}.
\newblock Pathtrack: Fast trajectory annotation with path supervision.
\newblock In {\em International Conference on Computer Vision (ICCV)}, 2017.

\bibitem{cvpr2015}
Ishan Misra, Abhinav Shrivastava, and Martial Hebert.
\newblock Watch and learn: Semi-supervised learning of object detectors from
  videos.
\newblock {\em CoRR}, 2015.

\bibitem{trackingnet}
Matthias M{\"{u}}ller, Adel Bibi, Silvio Giancola, Salman Al{-}Subaihi, and
  Bernard Ghanem.
\newblock Trackingnet: {A} large-scale dataset and benchmark for object
  tracking in the wild.
\newblock {\em ECCV}, 2018.

\bibitem{davis}
F. Perazzi, J. Pont-Tuset, B. McWilliams, L. {Van Gool}, M. Gross, and A.
  Sorkine-Hornung.
\newblock A benchmark dataset and evaluation methodology for video object
  segmentation.
\newblock In {\em Computer Vision and Pattern Recognition}, 2016.

\bibitem{qi2017pointnet}
Charles~R Qi, Hao Su, Kaichun Mo, and Leonidas~J Guibas.
\newblock Pointnet: Deep learning on point sets for 3d classification and
  segmentation.
\newblock {\em Proc. Computer Vision and Pattern Recognition (CVPR), IEEE},
  2017.

\bibitem{momentum}
Ning Qian.
\newblock On the momentum term in gradient descent learning algorithms.
\newblock {\em Neural Networks}, Jan. 1999.

\bibitem{ytbb}
Esteban Real, Jonathon Shlens, Stefano Mazzocchi, Xin Pan, and Vincent
  Vanhoucke.
\newblock Youtube-boundingboxes: {A} large high-precision human-annotated data
  set for object detection in video.
\newblock {\em CoRR}, abs/1702.00824, 2017.

\bibitem{noisylabels}
David Rolnick, Andreas Veit, Serge~J. Belongie, and Nir Shavit.
\newblock Deep learning is robust to massive label noise.
\newblock {\em CoRR}, 2017.

\bibitem{ImageNetpaper}
Olga Russakovsky, Jia Deng, Hao Su, Jonathan Krause, Sanjeev Satheesh, Sean Ma,
  Zhiheng Huang, Andrej Karpathy, Aditya Khosla, Michael Bernstein,
  Alexander~C. Berg, and Li Fei-Fei.
\newblock Imagenet large scale visual recognition challenge.
\newblock {\em Int. J. Comput. Vision}, 115(3):211--252, Dec. 2015.

\bibitem{imagenetvid}
Olga Russakovsky, Jia Deng, Hao Su, Jonathan Krause, Sanjeev Satheesh, Sean Ma,
  Zhiheng Huang, Andrej Karpathy, Aditya Khosla, Michael Bernstein,
  Alexander~C. Berg, and Li Fei-Fei.
\newblock {ImageNet Large Scale Visual Recognition Challenge}.
\newblock {\em International Journal of Computer Vision (IJCV)},
  115(3):211--252, 2015.

\bibitem{objects365}
Shuai Shao, Zeming Li, Tianyuan Zhang, Chao Peng, Gang Yu, Xiangyu Zhang, Jing
  Li, and Jian Sun.
\newblock Objects365: A large-scale, high-quality dataset for object detection.
\newblock In {\em The IEEE International Conference on Computer Vision (ICCV)},
  October 2019.

\bibitem{waymoopendataset}
Pei Sun, Henrik Kretzschmar, Xerxes Dotiwalla, Aurelien Chouard, Vijaysai
  Patnaik, Paul Tsui, James Guo, Yin Zhou, Yuning Chai, Benjamin Caine, Vijay
  Vasudevan, Wei Han, Jiquan Ngiam, Hang Zhao, Aleksei Timofeev, Scott
  Ettinger, Maxim Krivokon, Amy Gao, Aditya Joshi, Yu Zhang, Jonathon Shlens,
  Zhifeng Chen, and Dragomir Anguelov.
\newblock Scalability in perception for autonomous driving: Waymo open dataset,
  2019.

\bibitem{oxuva}
Jack Valmadre, Luca Bertinetto, Jo{\~{a}}o~F. Henriques, Ran Tao, Andrea
  Vedaldi, Arnold W.~M. Smeulders, Philip H.~S. Torr, and Efstratios Gavves.
\newblock Long-term tracking in the wild: {A} benchmark.
\newblock {\em ECCV}, 2018.

\bibitem{alprop}
Sudheendra Vijayanarasimhan and Kristen Grauman.
\newblock Active frame selection for label propagation in videos.
\newblock pages 496--509, 10 2012.

\bibitem{eccv12ALGrauman}
Sudheendra Vijayanarasimhan and Kristen Grauman.
\newblock Active frame selection for label propagation in videos.
\newblock In {\em ECCV 2012}, 2012.

\bibitem{vatical}
Carl Vondrick and Deva Ramanan.
\newblock Video annotation and tracking with active learning.
\newblock In J. Shawe-Taylor, R.~S. Zemel, P.~L. Bartlett, F. Pereira, and
  K.~Q. Weinberger, editors, {\em Advances in Neural Information Processing
  Systems 24}, pages 28--36. Curran Associates, Inc., 2011.

\bibitem{vatic}
Carl Vondrick, Deva Ramanan, and Donald Patterson.
\newblock Efficiently scaling up video annotation with crowdsourced
  marketplaces.
\newblock In {\em ECCV}, ECCV'10, pages 610--623, Berlin, Heidelberg, 2010.
  Springer-Verlag.

\bibitem{labelme2016}
Ketaro Wada.
\newblock {labelme: Image Polygonal Annotation with Python}.
\newblock \url{https://github.com/wkentaro/labelme}, 2016.

\bibitem{siammask}
Qiang Wang, Li Zhang, Luca Bertinetto, Weiming Hu, and Philip~HS Torr.
\newblock Fast online object tracking and segmentation: A unifying approach.
\newblock {\em arXiv preprint arXiv:1812.05050}, 2018.

\bibitem{supertrajectory}
Wenguan Wang and Shenjian Bing.
\newblock Super-trajectory for video segmentation.
\newblock {\em ICCV}, 2017.

\bibitem{youtbevos}
Ning Xu, Linjie Yang, Yuchen Fan, Dingcheng Yue, Yuchen Liang, Jianchao Yang,
  and Thomas~S. Huang.
\newblock Youtube-vos: {A} large-scale video object segmentation benchmark.
\newblock {\em CoRR}, 2018.

\bibitem{ladcf}
Tianyang Xu, Zhen{-}Hua Feng, Xiao{-}Jun Wu, and Josef Kittler.
\newblock Learning adaptive discriminative correlation filters via temporal
  consistency preserving spatial feature selection for robust visual tracking.
\newblock {\em CoRR}, 2018.

\bibitem{mbmd}
Yunhua Zhang, Dong Wang, Lijun Wang, Jinqing Qi, and Huchuan Lu.
\newblock Learning regression and verification networks for long-term visual
  tracking.
\newblock {\em CoRR}, 2018.

\bibitem{DaSiamRPN}
Zheng Zhu, Qiang Wang, Li Bo, Wei Wu, Junjie Yan, and Weiming Hu.
\newblock Distractor-aware siamese networks for visual object tracking.
\newblock In {\em European Conference on Computer Vision}, 2018.

\end{thebibliography}
}

\end{document}